\pdfoutput=1
\documentclass[11pt]{article}
\usepackage{ACL2023}
\usepackage{times}
\usepackage{booktabs}
\usepackage{tabularx}

\newcommand\clearrow{\global\let\rowmac\relax}
\usepackage{multirow}
\usepackage{graphicx}
\usepackage{tikz}
\usepackage{amssymb}
\usepackage{pifont}
\usepackage{todonotes}
\usepackage{xcolor}
\usepackage{soul}
\sethlcolor{yellow}

\newcommand{\cmark}{\ding{51}}%
\newcommand{\xmark}{\ding{55}}%

\newcommand{\chatgpt}{\textsc{ChatGPT}}
\newcommand{\chatcompanion}{\textsc{LLM-Augmenter}\xspace}
\usepackage{tikz}

\newcommand{\lmcompanion}{{\sc LLM-Augmenter}\xspace}
\usepackage{amsmath}
\usepackage{amsfonts}
\usepackage{amssymb}
\usepackage{wrapfig}
\usepackage{subcaption}
\usepackage{multirow}
 \usepackage{mathtools} 

\usepackage{verbatim}

\usepackage{anyfontsize}

\usepackage{color}
\usepackage{tikz}
\usetikzlibrary{arrows,shapes,snakes,automata,backgrounds,fit,petri}
\usepackage{adjustbox}

\newcommand{\RN}[1]{%
	\textup{\lowercase\expandafter{\it \romannumeral#1}}%
}

\usepackage{csquotes}
\usepackage{booktabs}

\usepackage{tcolorbox}
\usepackage{colortbl}

\usepackage{xcolor}
\definecolor{mygreen}{HTML}{3cb44b}
\colorlet{myyellow}{green!10!orange!90!}
\makeatletter
\newcommand{\distas}[1]{\mathbin{\overset{#1}{\kern\z@\sim}}}%

\newcommand*\circled[1]{\tikz[baseline=(char.base)]{%
            \node[shape=circle,fill=blue!10,draw,inner sep=1.5pt] (char) {#1};}}

\usepackage{enumitem}

\usepackage[lined,boxed,commentsnumbered,ruled,linesnumbered]{algorithm2e}

\newcommand{\ie}[0]{\emph{i.e., }}

\newcommand{\eg}[0]{\emph{e.g., }}

\newcommand{\beq}{\vspace{0mm}\begin{equation}}
\newcommand{\eeq}{\vspace{0mm}\end{equation}}
\newcommand{\beqs}{\vspace{0mm}\begin{eqnarray}}
\newcommand{\eeqs}{\vspace{0mm}\end{eqnarray}}
\newcommand{\barr}{\begin{array}}
\newcommand{\earr}{\end{array}}

\ifx\assumption\undefined
\newtheorem{assumption}{Assumption}
\fi

\ifx\definition\undefined
\newtheorem{definition}{Definition}
\fi

\ifx\remark\undefined
\newtheorem{remark}{Remark}
\fi

\usepackage{amsmath}
\DeclareMathOperator*{\argmax}{argmax}

\title{Check Your Facts and Try Again: Improving Large Language Models with External Knowledge and Automated Feedback\thanks{~~Correspondence: \{bapeng,mgalley,jfgao\}microsoft.com}}

\author{
Baolin Peng\textsuperscript{$\dagger$} \hspace{.1cm} 
\textbf{Michel Galley}\textsuperscript{$\dagger$} \hspace{.1cm}
\textbf{Pengcheng He}\textsuperscript{$\dagger$} \hspace{.1cm}
\textbf{Hao Cheng}\textsuperscript{$\dagger$} \hspace{.1cm}
\textbf{Yujia Xie}\textsuperscript{$\dagger$} \hspace{.1cm}\\
\textbf{Yu Hu}\textsuperscript{$\dagger$}
\textbf{Qiuyuan Huang}\textsuperscript{$\dagger$} 
\textbf{Lars Liden}\textsuperscript{$\dagger$} 
\textbf{Zhou Yu}\textsuperscript{$\ddagger$} 
\textbf{Weizhu Chen}\textsuperscript{$\dagger$} 
\textbf{Jianfeng Gao}\textsuperscript{$\dagger$} \\
\textsuperscript{$\dagger$} Microsoft Research \hspace{.1cm} \textsuperscript{$\ddagger$} Columbia University\\
}

\begin{document}
\maketitle
\begin{abstract}
Large language models (LLMs), such as ChatGPT, are able to generate human-like, fluent responses for many downstream tasks, \eg task-oriented dialog and question answering.
However, applying LLMs to real-world, mission-critical applications remains challenging mainly due to their tendency to generate hallucinations and their inability to use external knowledge.
This paper proposes a \lmcompanion system, which augments a black-box LLM with a set of plug-and-play modules. Our system makes the LLM generate responses grounded in external knowledge, \eg stored in task-specific databases. It also iteratively revises LLM prompts to improve model responses using feedback generated by utility functions, \eg the factuality score of a LLM-generated response.
The effectiveness of \lmcompanion is empirically validated on two types of scenarios, task-oriented dialog and open-domain question answering. \lmcompanion significantly reduces ChatGPT's hallucinations without sacrificing the fluency and informativeness of its responses. 
We make the source code and models publicly available.\footnote{\url{https://aka.ms/llm-augmenter}}

\end{abstract}

\section{Introduction}
\label{sec:intro}
Large Language models (LLMs), such as \mbox{GPT-3} \cite{NEURIPS2020_1457c0d6} and ChatGPT,  
have demonstrated an outstanding ability in generating fluent, coherent, and informative natural language texts. 
It is commonly understood that the impressive capabilities of these models stem from the abundance of world knowledge encoded therein and models' ability to generalize from that knowledge. 
However, the knowledge encoding of LLMs is lossy and the knowledge generalization could lead to ``memory distortion.'' 
As a result, these models tend to \emph{hallucinate}, which can cause damage when deployed for mission-critical tasks. 
In addition, even with exponentially growing model sizes, LLMs can never encode all information needed for many applications. For example, constant changes in real-world settings cause LLMs to quickly become stale for time-sensitive tasks such as news question answering, and many proprietary datasets are not available for LLM training due to privacy.
While there is a growing interest in improving LLMs using external knowledge \cite[\eg][]{ghazvininejad2017knowledge,Guu2020REALMRL,Zhong2022TrainingLM,GaoEtAl:2019,gao2022neural},
almost all the previously proposed methods require finetuning the parameters of a LLM, %
which can be prohibitively expensive as the size of LLMs grows exponentially.
Thus, it is highly desirable to augment a \emph{fixed} LLM with plug-and-play (PnP) modules for mission-critical tasks.

\begin{figure}
\centering
\includegraphics[width=1.0\columnwidth]
{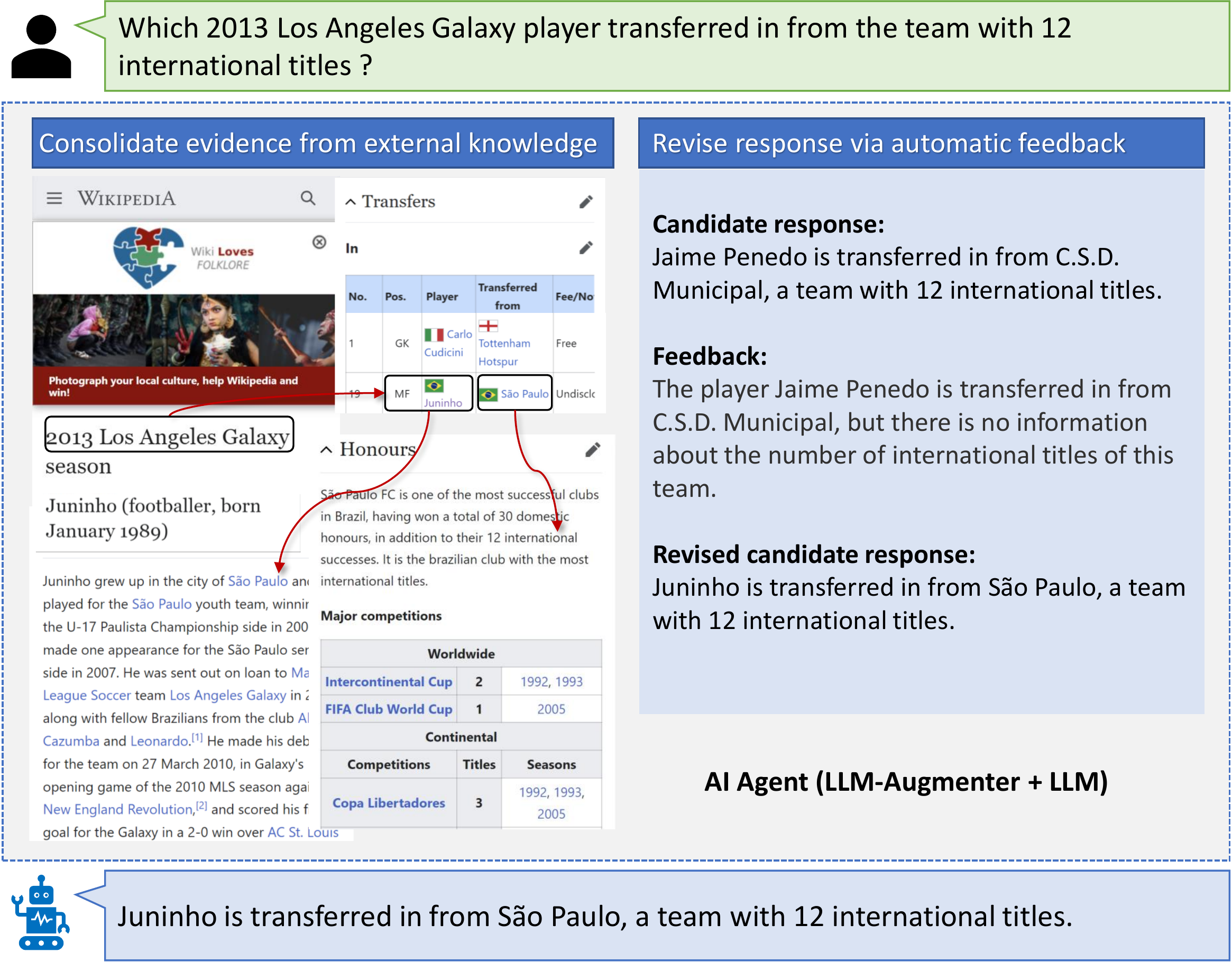}
\caption{\lmcompanion improves a fixed LLM by (1) consolidating evidence from external knowledge for the LLM to generate responses grounded in evidence, and (2) revising LLM's (candidate) responses using automated feedback.} 
\label{fig:motivation}
\end{figure} 

In this paper, we present \lmcompanion to improve LLMs with external knowledge and automated feedback using PnP modules.
As illustrated by the example in Figure 1, given a user query (\eg regarding a 2013 Los Angeles Galaxy player transfer), \lmcompanion first retrieves evidence from external knowledge (\eg Web or task-specific datasets) and, if necessary, further consolidates evidence by linking retrieved raw evidence with related context 
(\eg information of the entity ``2013 Los Angeles Galaxy'')
and performing reasoning to form evidence chains (\eg table-passage in the figure).
Then, \lmcompanion queries a fixed LLM (\ie ChatGPT in our study) using a prompt that contains the consolidated evidence for ChatGPT to generate a candidate response grounded in external knowledge (evidence).
\lmcompanion then verifies the candidate response \eg by checking whether it hallucinates evidence. If so, \lmcompanion generates a feedback message (\eg about the team ``C.S.D. Municipal''). The message is used to revise the prompt to query ChatGPT again.  
The process iterates until a candidate response passes the verification and is sent to the user.  

In addition to proposing \lmcompanion, to be detailed in Section~\ref{sec:model}, we make the following contributions.
We perform an empirical study to validate the effectiveness of \lmcompanion using two tasks, information seeking dialog (Section 3) and open-domain Wiki question answering (Wiki QA) (Section 4). 
The study shows that \lmcompanion significantly reduces ChatGPT's hallucinations without sacrificing the fluency and informativeness of its generated responses.
For example, on the dialog task of customer service, human evaluation shows \lmcompanion improve ChatGPT by 32.3\% in Usefulness (measuring the groundedness or hallucination of model responses) and 12.9\% in Humanness (measuring the fluency and informativeness of model responses). 
The Wiki QA task is extremely challenging to ChatGPT in that answering these questions often requires multi-hop reasoning to piece together information of various modalities scattered across different documents. Our results show that although the \emph{closed-book} ChatGPT performs poorly and often hallucinates, \lmcompanion substantially improves the factuality score of the answers (absolute +10\% in F1) by grounding ChatGPT's responses in consolidated external knowledge and automated feedback.

\section{\lmcompanion}
\label{sec:model}
\begin{figure}
\centering
\includegraphics[width=1.0\columnwidth]
{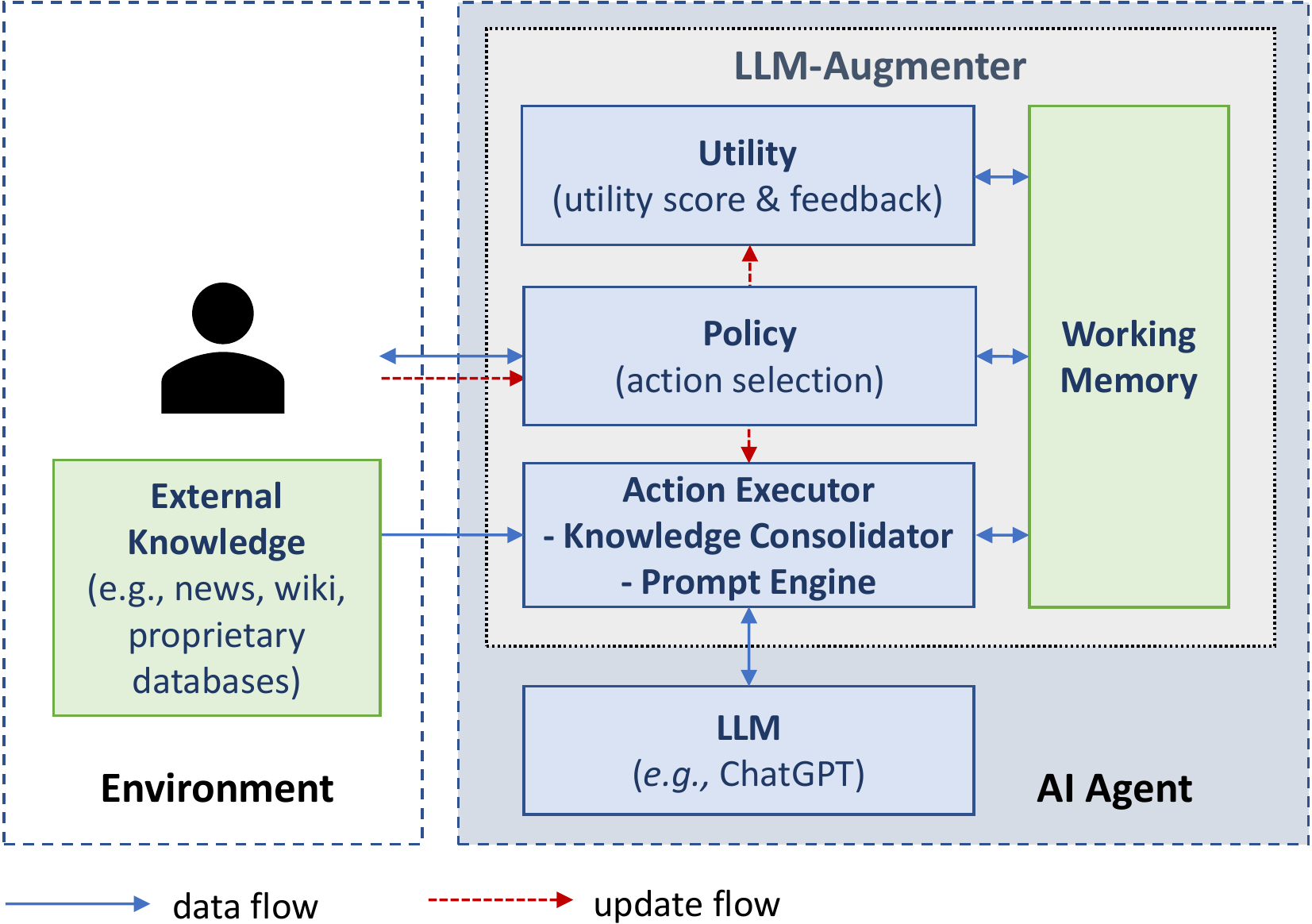}

\caption{\lmcompanion{} architecture showing how its plug-and-play modules interact with the LLM and the user's environment.}%
\label{fig:llm-augmenter-arch}
\end{figure}

The architecture of \lmcompanion is illustrated in Figure \ref{fig:llm-augmenter-arch}. It consists of a set of PnP modules (\ie Working Memory, Policy, Action Executor, and Utility) to improve a fixed LLM (\eg ChatGPT) with external knowledge and automated feedback to mitigate generation problems such as hallucination.

We formulate human-system conversation as a Markov Decision Process (MDP) described by a five-tuple $(\mathcal{S}, \mathcal{A}, {P}, {R}, \gamma)$:
\begin{itemize}%
    \item $\mathcal{S}$ is an infinite set of dialog states, which encode information stored in Working Memory, including dialog history, user query, evidence, candidate response; %
    \item $\mathcal{A}$ is a set of actions that Policy picks to execute, including (1) calling Knowledge Consolidator to consolidate evidence from external knowledge and (2) calling Prompt Engine to query the LLM to generate candidate responses; 
    \item ${P(s'|s,a)}$ gives the transition probability of entering a new state $s'$ after action $a$ is taken in state $s$;
    \item ${R(s,a)}$ is the external reward received after taking action $a$ in state $s$, which is provided by the environment (\eg users or simulators); and %
    \item $\gamma \in (0,1]$ is a discount factor.
\end{itemize}
In what follows, we describe in detail the modules of \lmcompanion.

\subsection{Working Memory}
This module tracks the dialog state that captures all essential information in the conversation so far. The state is represented using a six-tuple $(q, e, o, u, f, h_q)$:
\begin{itemize}%
    \item $q$ is the current user query;
    \item $e$ is evidence for $q$, consolidated from external knowledge by Knowledge Consolidator; 
    \item $o$ is a set of the LLM-generated candidate responses for $q$;
    \item $u$ is a score assessing the utility of each element of $o$, and $f$ is a verbalized feedback to guide the LLM to improve its utility --- both $u$ and $f$ are generated by the Utility module (see Section~\ref{sec:utility}); and
    \item $h_q$ is the dialog history before $q$.
\end{itemize}

Note that given user query $q$, \lmcompanion can take multiple iterations to revise its response, with each iteration generating a candidate response based on evidence, feedback and utility, before sending the final response to the user, as illustrated in Figure~\ref{fig:motivation}.

\subsection{Policy}

This module selects the next system action that leads to the best expected reward $R$. These actions include (1) acquiring evidence $e$ for $q$ from external knowledge, (2) calling the LLM to generate a candidate response, and (3) sending a response to users if it passes the verification by the Utility module.

The policy can be implemented using manually crafted rules, or trained on human-system interactions. In this study, we implement a trainable policy $\pi$ as a neural network model parameterized by $\theta$. $\pi_\theta$ is optimized using REINFORCE \citep{Williams1992SimpleSG} to maximize the expected reward as:

\begin{equation}
\argmax_\theta \mathbb{E}_{s \sim \mathcal{S}, a \sim \pi_\theta} [ {R}(s, a)]
\label{eq:policy}
\end{equation}

We find it effective to implement $\pi$ using a pre-trained model (e.g., T5), which allows us to not only leverage the capacity of the pre-trained model, but also to incorporate additional information through finetuning. 

Policy learning typically requires large amounts of human-machine interactions, which can be costly to collect. To address the challenge, policy learning can
be done in three stages:

\begin{itemize}
    \item Bootstrapping from a rule-based policy: Domain experts encode task-specific knowledge and business logic into IF-THEN rules. For example, if a product name is mentioned in a user query for customer service, it is wise to always call Knowledge Consolidator to collect information of the product from a product database.
    \item Learning with user simulators: We use a language model to simulate how human users interact with \lmcompanion{}. Any valid response from \lmcompanion{} that passes the evaluation of the Utility module can be used as a training example, allowing \lmcompanion{} to self-improve.
    \item Finally, \lmcompanion{} interacts with human users to further refine its policy.
\end{itemize}

In addition to Policy, the other trainable modules of \lmcompanion{} (\ie Knowledge Consolidator and Utility) can also be optimized using the same learning method.

\subsection{Action Executor}
This module performs an action selected by the policy. It is composed of two components, Knowledge Consolidator and Prompt Engine.

\subsubsection{Knowledge Consolidator} 

The Knowledge Consolidator augments LLMs with the capability of grounding their responses on external knowledge to mitigate hallucination when completing tasks, such as answering questions regarding latest news, and booking a table in a restaurant.
Following \cite{ma2022open}, the Knowledge Consolidator is designed in a modular fashion, consisting of a knowledge \emph{retriever}, an entity \emph{linker} and, an evidence \emph{chainer}. 

Specifically, the retriever first generates a set of search queries based on $q$ and $h_q$, and then calls a set of APIs to retrieve raw evidence from various external knowledge sources, such as calling Bing Search APIs to query Web documents including Wiki articles and Reddit messages, and REST APIs to query task-specific databases for restaurant reviews and product specifications.  

The retrieved raw evidence is sometimes incomplete and noisy. Thus, the entity linker enriches raw evidence with related context to form evidence graphs, \ie linking each entity mentioned in raw evidence to its corresponding description based on Wikipedia. 
Then, the chainer prunes irrelevant evidence from the graphs and forms a shortlist of evidence chains that are most relevant to queries.  
The consolidated evidence $e$ is then sent to Working Memory.
Figure~\ref{fig:motivation} shows an example of consolidated evidence for the anchored club ``Los Angeles Galaxy'', \ie two evidence chains corresponding to the transfer players in 2013 season and the former clubs, respectively.
 
\subsubsection{Prompt Engine}

The Prompt Engine generates a prompt 
to query the LLM to generate a (candidate) response $o$ for $q$. The prompt is a text string that consists of task instruction, user query $q$, dialog history $h_q$, evidence $e$ if it is made available by Knowledge Consolidator, and feedback $f$ if it is made available by the Utility module. Prompts are task-specific, and details thereof are provided in Appendix~\ref{sec:appendix}. %

\subsection{Utility}
\label{sec:utility}

Given a candidate response $o$, the Utility module generates utility score $u$ and a corresponding feedback $f$ using a set of task-specific utility functions. 

These utility functions\footnote{
Our experiments are with a single utility function. To allow multiple utility functions, we could learn a linear function mapping the outputs of these multiple functions to a single score using a linear function trained together with the other parameters of the policy.}
access the alignment of the LLM's responses with user expectations or specific business requirements. For example, in an information seeking dialog, it is important that all LLM's responses are preciously grounded in external evidence to avoid generating misleading or inaccurate information. In a restaurant reservation dialog, the LLM responses should be conversational and focused on guiding the user through the reservation process, rather than engaging in off-topic chitchats.

Inspired by \citet{sparrow}, there can be two distinct types of utility functions:
\begin{itemize}
    \item Model-based utility functions assign preference scores to different dimensions of a response, such as fluency, informativeness and factuality. These functions are trained on pre-collected human preference data or annotated log data.
    \item Rule-based utility functions, implemented using heuristics or programmed functions, measure whether a response complies with a specific rule.
\end{itemize}

In addition, we have developed a utility function to generate informative and actionable feedback to help revise prompts to allow the LLM to generate better responses. As shown in Figure~\ref{fig:motivation}, the utility function generates feedback ``but there is no information about the number of international titles.'' Such a utility function is a text generation model $Q$ parameterized by $\psi$, and can be implemented as a seq2seq or auto-regression language model. It tasks as input user query $q$, evidence $e$, candidate response $o$ and dialog history $h_q$, and generates feedback in text $f$ as 
\begin{equation}
f = Q_\psi(q, e, o, h_q)
\end{equation}
Alternatively, LLMs and rule-based natural language generator can be used for feedback generation.

In the next two sections, we present our experiments to validate the effectiveness of \lmcompanion in two types of distinct scenarios: (1) information seeking dialog, where the AI agent needs to generate informative and trustworthy responses based on a variety of external sources of knowledge, and (2) Wiki question answering, where the AI agent needs to answer questions by piecing together information of various modalities scattered among multiple Wiki documents.

\section{Information Seeking Dialog}
\label{sec:exp_kgr}
\subsection{Datasets}
\begin{table*}
	\centering
	\scalebox{0.74}{
	\setlength{\tabcolsep}{1.0mm}{
	\centering
		\begin{tabular}{lccrrrrrrrrr}
			\toprule
			   Model & K.C. & Feedback &
                    $\mathtt{KF1} \uparrow$ & $\mathtt{BLEU} \uparrow$ & $\mathtt{ROUGE} \uparrow$ & $\mathtt{chrF} \uparrow$ & $\mathtt{METEOR} \uparrow$ & $\mathtt{BERTScore} \uparrow$ & $\mathtt{BARTScore} \uparrow$ & $\mathtt{BLEURT} \uparrow$ & Avg. length\\
\midrule
\chatgpt{}       & -    & -      & 26.71 & 1.01 & 16.78 & 23.80 & 7.34 & 82.14 & 0.25 & 26.98 & 58.94\\
\midrule
\chatcompanion{} & BM25 & \xmark & 34.96 & 6.71 & 22.25 & 27.02 & 9.35 & {\bf 83.46} & 0.34 & 26.89 & 46.74\\
\chatcompanion{} & BM25 & \cmark & {\bf 36.41} & {\bf 7.63} & {\bf 22.80} & {\bf 28.66} & {\bf 10.17} & 83.33 & {\bf 0.35} & {\bf 27.71} & 54.24\\
\midrule
\chatcompanion{} & gold & \xmark & {\it 57.44} & {\it 19.24} & {\it 38.89} & {\it 40.02} & {\it 17.21} & {\it 86.65} & {\it 0.82} & {\it 40.55} & {\it 44.35}\\
\chatcompanion{} & gold & \cmark & {\it 60.76} & {\it 21.49} & {\it 40.56} & {\it 42.14} & {\it 18.50} & {\it 86.89} & {\it 0.93} & {\it 42.15} & {\it 47.19}\\
\bottomrule
		\end{tabular}
	}}
		    
	\caption{Evaluation scores (in \%) and average response lengths for the News Chat (DSTC7) dataset. BM25: Each model retrieves 5 knowledge snippets from the corresponding knowledge source. K.C. denotes Knowledge Consolidator.}
	\label{tab:results_dstc7}
\end{table*}

\begin{table*}
	\centering
	\scalebox{0.74}{
	\setlength{\tabcolsep}{1.0mm}{
	\centering
		\begin{tabular}{lccrrrrrrrrr}
			\toprule
			   Model & K.C. & Feedback &
                    $\mathtt{KF1} \uparrow$ & $\mathtt{BLEU} \uparrow$ & $\mathtt{ROUGE} \uparrow$ & $\mathtt{chrF} \uparrow$ & $\mathtt{METEOR} \uparrow$ & $\mathtt{BERTScore} \uparrow$ & $\mathtt{BARTScore} \uparrow$ & $\mathtt{BLEURT} \uparrow$ & Avg. length\\
\midrule
\chatgpt{}       & -    & -      & 31.33 & 4.70 & 24.02 & 27.14 & 12.83 & 87.88 & 1.53 & {\bf 47.99} & 28.81\\
\midrule
\chatcompanion{} & BM25 & \xmark & 34.07 & {\bf 4.78} & {\bf 24.52} & 28.95 & 13.61 & {\bf 87.96} & 1.78 & 47.21 & 32.65\\
\chatcompanion{} & BM25 & \cmark & {\bf 37.41} & 3.86 & 24.20 & {\bf 30.90} & {\bf 14.74} & 87.58 & {\bf 2.09} & 44.71 & 45.07\\
\midrule
\chatcompanion{} & gold & \xmark & {\it 45.63} & {\it 6.54} & {\it 29.77} & {\it 33.32} & {\it 16.93} & {\it 89.35} & {\it 2.59} & {\it 54.38} & {\it 33.04}\\
\chatcompanion{} & gold & \cmark & {\it 52.83} & {\it 5.63} & {\it 29.65} & {\it 35.68} & {\it 18.66} & {\it 89.01} & {\it 3.14} & {\it 52.49} & {\it 45.09}\\
\bottomrule
		\end{tabular}
	}}
	\caption{Evaluation scores (in \%) and average response lengths for the Customer Service (DSTC11) dataset. BM25: Each model retrieves 5 knowledge snippets from the corresponding knowledge source. K.C. denotes Knowledge Consolidator.}
	\label{tab:results_dstc11}
\end{table*}
\paragraph{News Chat:} We repurpose the DSTC7 Track 2 task as an evaluation corpus for news conversation. The goal of this task is to generate informative responses that are grounded in external knowledge (\ie news) and go beyond chitchat. We followed the data crawling process used in DSTC7 Task 2~\cite{dstc7task2}. We started by selecting Reddit discussion threads that contained URLs in the description, which were crawled from various news-related subreddits during the time period of 2021-2022. We then restricted the URL domain to a curated list of news websites, and extracted the relevant oracle passage by selecting the most appropriate passage for the context based on ROUGE-F1 scores~\cite{lin2004rouge}. In order to reduce noisy or irrelevant information, we only kept examples with an F1 score higher than a certain threshold, resulting in a total of 1370 examples for evaluation. 
\paragraph{Customer Service:} We use DSTC11 Track 5 \cite{kim2023} as a showcase in a conversational customer service scenario. It expands upon the DSTC9 Track 1 dataset by incorporating subjective knowledge from customer reviews in addition to factual knowledge from FAQs. This allows users to have an engaging and informative conversational experience with the AI system. The dataset evaluates the ability of the AI agent to understand relevant user review posts and FAQs, and generate responses based on both reviews and FAQ snippets. It is collected based on the MultiWOZ 2.1~\cite{eric2019multiwoz} dataset and includes users' knowledge-seeking queries that require the AI agent to use FAQs and user reviews to respond. There are 14768 dialog sessions for training and validation, and the test set is currently unavailable. Therefore, we used the validation set for our evaluations. 

\subsection{Experiment Setup}

\paragraph{Language Model:} Throughout this work, we focus on using ChatGPT as the backbone black-box LLM.
It is straightforward to apply \lmcompanion to other LLMs, such as GPT-3~\cite{NEURIPS2020_1457c0d6} or PaLM~\cite{chowdhery2022palm}.

\paragraph{Knowledge Consolidator:} For News Chat, Knowledge Consolidator includes a BM25 retriever over web documents linked from Reddit posts. For the Customer Service task, Knowledge Consolidator includes a BM25-based retriever over the knowledge bases of FAQs and Yelp reviews. 

Additionally, we also experiment with ground-truth knowledge, referred to as \emph{golden knowledge} henceforth, which is used by human annotators during data collection, in our oracle experiments. 

\paragraph{Prompt Engine:} The prompt templates utilized for News Chat and Customer Service are shown in the appendix in Table \ref{tab:appendix_prompt_templates_news_chat} and Table \ref{tab:appendix_prompt_templates_customer_service}, respectively.

\paragraph{Utility:} The goal of this task is to generate responses that are coherent to the context and grounded in external knowledge. To evaluate the degree to which the generated responses are grounded in consolidated evidence, we use the utility score, Knowledge F1 \cite{shuster2021retrieval}, to measure the overlap between a prediction and evidence which is either consolidated by Knowledge Consolidator or provided as golden knowledge. Feedback generation is accomplished using a template-based natural language generator.\footnote{If the KF1 score falls below a certain threshold, the feedback is ``The response is inconsistent with the knowledge. Please generate again.''} In addition, we use ChatGPT as a utility function, \ie self-criticism to gather feedback by prompting ChatGPT to evaluate candidate responses and give feedback on how to improve them.

\paragraph{Policy:} 
Due to ChatGPT's current limited bandwidth, we use a rule-based policy for our experiments involving ChatGPT.
The prior knowledge about this task inspired us to design a policy that always uses Knowledge Consolidator, evaluates the quality of a candidate response using ChatGPT, and provides feedback to revise the prompt. 
Additionally, to test the viability of \lmcompanion{} with a trainable policy, we employ offline RL to train the parameters of Policy as Equation~\ref{eq:policy}, where the policy model is based on T5-Base.

\paragraph{Evaluation:} We evaluate the performance of \lmcompanion{} on information-seeking dialog tasks using both automatic metrics and human evaluations. 
Following the literature, we consider commonly used metrics, Knowledge F1 (KF1) and BLEU-4, in grounded conversational response generation and task-oriented dialog. BLEU \cite{papineni2002bleu} measures the overlap between the model's output and the ground-truth human response, while KF1 \cite{Lian2019LearningTS} assesses the overlap with the knowledge that the human used as a reference during dataset collection.
Additionally, we include ROUGE-1 \cite{lin2004rouge} and METEOR \cite{banerjee2005meteor} as these metrics have been found to best correlate with human judgment on the DSTC9 and DSTC11 customer support tasks \cite{kim2020beyond}.
We further include BLEURT \cite{sellam-etal-2020-bleurt}, BERTScore \cite{zhang2019}, chrF \cite{popovic-2015-chrf}, which have been shown to be among the best-performing text generation metrics on dialog\cite{yeh2021,Peng2022GODELLP}.
Lastly, we also consider BARTScore as it has been reported to be one of the best model-based metrics \cite{yuan2021}. Given that BARTScore can be interpreted as a log-probability, we report results with its natural exponent (positive scores).  
Additionally, we perform a turn-level human evaluation to investigate whether responses are (1) useful and (2) human-like. Following the evaluation protocol by \cite{Peng2022GODELLP}, using Amazon Mechanical Turk, we hired master-level workers with lifetime HIT acceptance rate above 95\%, and asked them to answer two questions on usefulness (\ie \textit{which response sounds more useful}) and humanness (\ie \textit{which speaker sounds more human}).

\subsection{Automatic Evaluation Results}

\paragraph{The impact of using external knowledge:} Experiment results are shown in Tables \ref{tab:results_dstc7} and \ref{tab:results_dstc11}. We observe that ChatGPT achieves reasonable performance even in the zero-shot setting. However, with access to golden knowledge, the performance is dramatically improved. This suggests that while LLMs are able to encode a large amount of general knowledge in their parameters, they can still benefit from more specific, targeted knowledge. This is likely because LLMs are designed to handle a wide range of tasks and therefore may not always have access to the most relevant or up-to-date information for a given task. 
Our experiments show that providing LLMs with task-specific knowledge can significantly mitigate hallucination without sacrificing the fluency and informativeness of model-generated responses. 
As demonstrated in Tables \ref{tab:results_dstc7} and \ref{tab:results_dstc11}, \lmcompanion{} mitigates ChatGPT's hallucination issue on both the news chat and customer service tasks. Specifically, we observe a significant improvement in KF1 scores of approximately 10 and 6 points, respectively, due to the use of evidence retrieved by Knowledge Consolidator.

\begin{figure}%
  \centering  
  \includegraphics[width=1.1\linewidth]{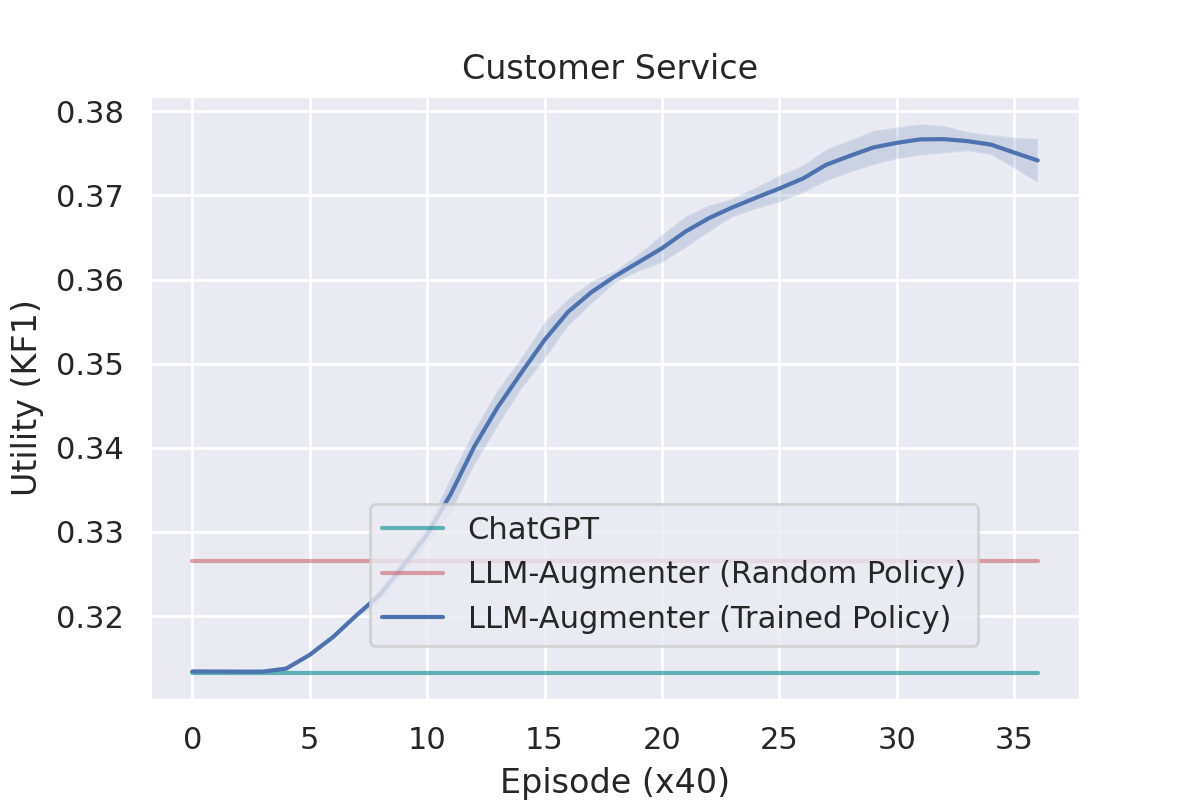}
  \caption{Learning curve of ChatGPT-Companion with T5-Base as the policy module. The solid curves are the mean and the shaded regions are the maximum and minimum utility scores over 5 runs.}
  \label{fig:llmscompanion_learning_curve}
\end{figure}
\vspace{-2mm}

\paragraph{The impact of using automated feedback:} As listed in Tables \ref{tab:results_dstc7} and \ref{tab:results_dstc11}, the results of using golden knowledge setting demonstrate that incorporating feedback from the Utility module leads to substantial improvement 3.3 points in KF1 on News Chat and 7.2 on Customer Service, respectively. Similarly, significant improvement can also be observed when using evidence provided by Knowledge Consolidator.

\begin{figure}%
  \centering  
  \includegraphics[width=1.1\linewidth]{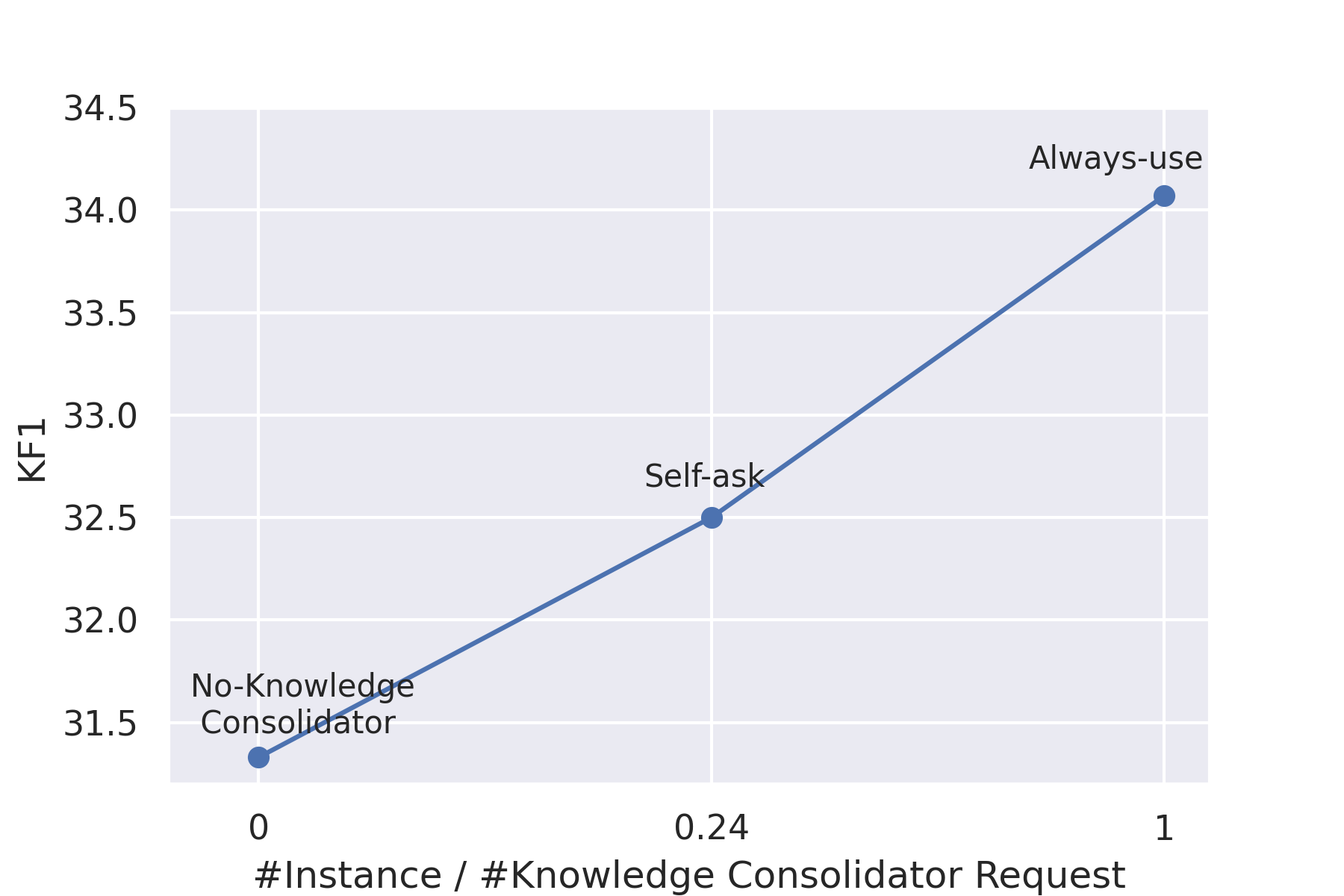}
  \caption{Ablation studies on different policies of \chatcompanion{} in Customer Service scenario.}
  \label{fig:ablation_plot_toolset}
\end{figure}

\paragraph{The impact of using trainable Policy:}
Figure \ref{fig:llmscompanion_learning_curve} shows the learning curve of \lmcompanion{} on the customer service task.
As we do not have an external reward that would require collecting data from real users, we instead define here our reward $R$ as the KF1 utility function. This helps demonstrate the effectiveness of \lmcompanion{} in its reinforcement learning (RL) setup. 
As our experiments are akin to single turn interactions, we did not need to set discount factor $\gamma$, but future work may need to rely on it.
We see that \lmcompanion{}'s reward on test data increases as the number of training episodes (dialog sessions) increases, surpassing a random policy after 600 interactions and ultimately reaching a KF1 score of approximately 37.5. Through these interactions, \lmcompanion{} is able to learn to effectively select the next system action to maximize the reward, which helps our system reduce hallucinations while generating fluent and informative responses. 

\begin{table}%
	\centering
		    
	\setlength{\tabcolsep}{1.0mm}{
 \scalebox{0.85}{
		\begin{tabular}{l|cc}
			\toprule
						    
			    Model             &  Usefulness $ \uparrow$ & Humanness $ \uparrow$ \\
			\midrule

 ChatGPT & 34.07 & 30.92 \\
  \chatcompanion{} & {\bf 45.07} & {\bf 35.22} \\
   \bottomrule
		\end{tabular}
	}
 }
		    
	\caption{Human evaluation of ChatGPT and \lmcompanion{} with BM25 in Customer Service scenario. All differences are significant (p $<$ 0.05). Inter-annotator agreements according Krippendorff’s alpha (interval metric) are 0.15 and 0.07 respectively.}
	\label{tab:results_human_evaluation}
\end{table}

\subsection{Human Evaluation Results}

We compare ChatGPT with and without \lmcompanion{}. A total of 948 randomly selected examples from the customer service dataset are used for human evaluation. The evaluation results are converted from a 5-point Likert-like scale to a win/tie/loss scale for reporting, as shown in Table \ref{tab:results_human_evaluation}. We observe a strong preference for \lmcompanion{} over ChatGPT alone in terms of both Usefulness and Humanness. The result is consistent with the automatic evaluation result, discussed earlier. 

\subsection{Ablation Study}

\begin{figure}%
  \centering  
  \includegraphics[width=1.1\linewidth]{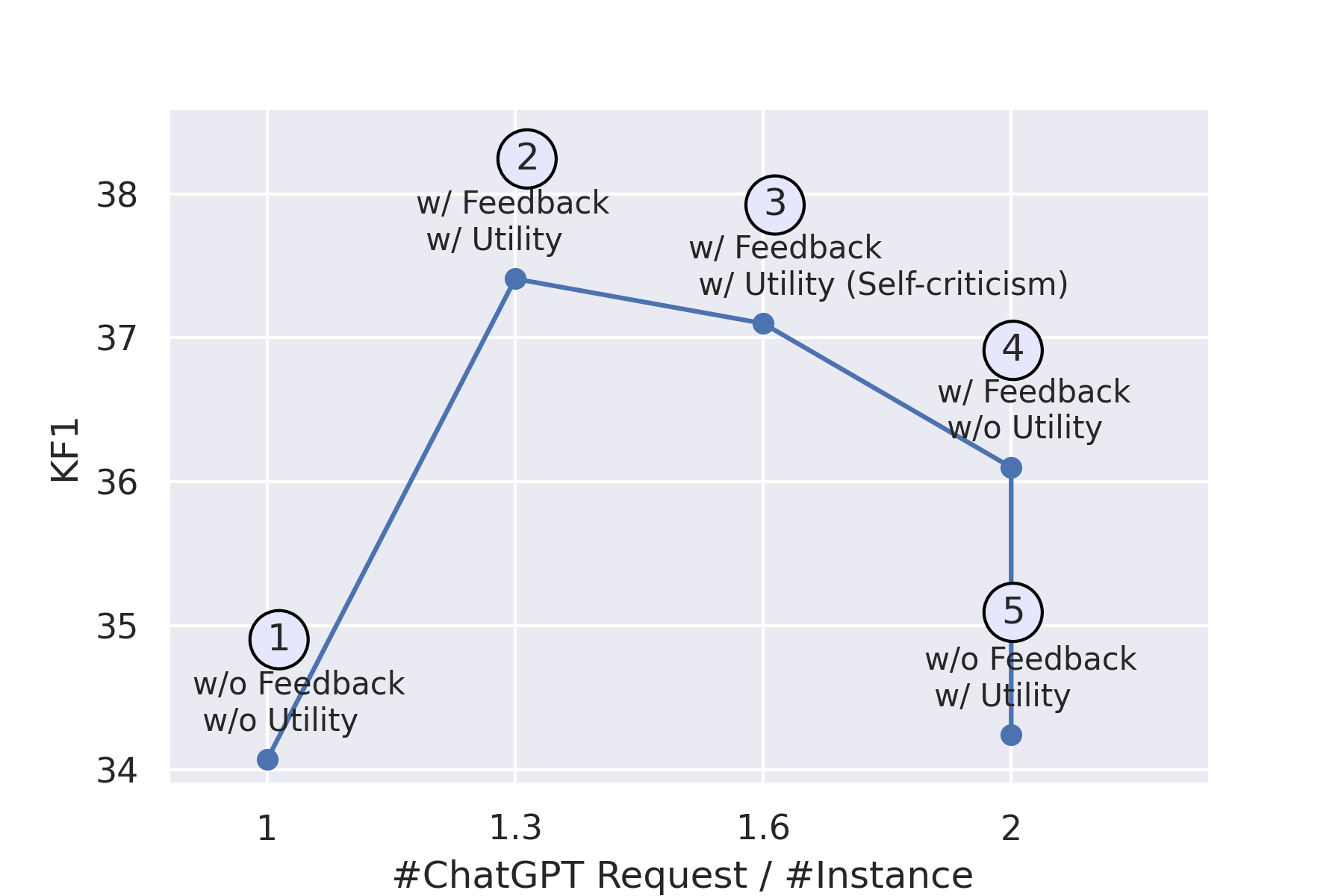}
  \caption{ \lmcompanion{} benefits from the combination of using utility function and iterative improvement with feedback. The x-axis indicates the average number of ChatGPT prompting and the y-axis is the KF1. The studies are conducted in the Customer Service scenario with knowledge being provided by BM25.}
  \label{fig:ablation_plot_feedback_utility}
\end{figure}

We conduct ablation experiments to evaluate the effect of various policies on the utilization of the knowledge consolidator. Figure \ref{fig:ablation_plot_toolset} shows the performance of three different variants of the policy: 1) no-knowledge consolidator, in which the knowledge consolidator is not used, 2) Self-ask, in which the knowledge consolidator is only utilized when the LM suggests the use of external knowledge by prompting it whether to use, and (3) Always-use, in which the knowledge consolidator is always provided to the LM. Our results indicate that Self-ask policy achieves a significantly better KF1 score than the No-knowledge consolidator policy, with the ChatGPT model unable to answer user queries and suggesting knowledge consolidator access for 24\% of examples. However, the Always-use policy, while achieving the best KF1 score, also incurred additional overhead in terms of knowledge consolidator access. These observations suggest that a trainable policy model should be employed to learn when to use external knowledge.

In addition, the evaluation results on the impact of different types of feedback for \chatcompanion{} are listed in Table \ref{tab:ablation_self_critic_results}. We observe that self-criticism feedback enhances response quality make it more knowledge-grounded. Although its performance is comparable to that of rule-based feedback, it provides more detailed suggestions. We speculate that self-criticism will be more helpful for complex tasks. Some examples can be found in \ref{tab:appendix_example}.

To understand the impact of utility functions and feedback-augmented prompting on the performance of \lmcompanion{}, we conduct an analysis by turning each component on and off. Figure \ref{fig:ablation_plot_feedback_utility} illustrates the results of each variant. We observe that the combination of using utility functions and feedback-augmented prompting, \ie \circled{2}, achieves the best performance. In addition, always providing feedback (as shown in \circled{4}) also enhances the performance, although it requires additional model prompting. \circled{5} represents prompting ChatGPT twice and re-ranking the response based on the utility functions, which results in a slightly higher KF1, but performs significantly worse than \circled{2}. These findings suggest that incorporating both utility functions and feedback is a more effective method for improving the alignment of LLMs.

\begin{table}%
	\centering
		    
	\setlength{\tabcolsep}{1.0mm}{
		\begin{tabular}{l|c}
			\toprule
						    
			    Feedback             &  $\mathtt{KF1} \uparrow$ \\
			\midrule
            w/o Feedback &  34.07 \\
            w/ Rule-based Feedback &  37.41 \\
            w/ Self-criticism Feedback &  37.10 \\

   \bottomrule
		\end{tabular}
	}
		    
	\caption{Ablation studies on different feedback of \chatcompanion{} in Customer Service scenario.}
	\label{tab:ablation_self_critic_results}
\end{table}

\section{Wiki QA}
Instead of conversational evaluations, we focus on stress tests on ChatGPT here using open-domain question answering.
As ChatGPT and other LLMs are mostly trained using abundant text from single web pages, we hypothesize that answering multi-hop questions involving scattered information across different pages/modalities can better serve the purpose.
Due to this, closed-book LLMs are more likely to hallucinate.
Moreover, the complex step-by-step reasoning can even be challenging for existing search systems to gather all necessary support evidence in one-shot.
Thus, more advanced knowledge consolidation techniques are essential to elicit LLMs for proper grounding.
Lastly, different from conversational tasks where long-form responses are desirable, we mainly consider questions with concise short-form answers, i.e., there exists a significant style shift in responses.
To align ChatGPT to this new scenario with distinct characteristics, extra instructions are needed.

\begin{table*}%
	\centering
		    
	\setlength{\tabcolsep}{1.0mm}{
		\begin{tabular}{lccccc}
			\toprule
						    
			\multirow{2}{*}{Model} & &  & \multicolumn{3}{c}{Wiki QA} \\
			\cmidrule(l){4-6}  
			                 & Knowledge Consolidator & Feedback  & $\mathtt{P} \uparrow$ & $\mathtt{R} \uparrow$ & $\mathtt{F1} \uparrow$ \\
			\midrule
			\chatgpt{}       & -       & -                          & 0.48                  & 1.52                  & 0.59                   \\
			\midrule
			\chatcompanion{} & DPR  & \xmark   & 2.08                    & 4.31                     & 2.38                      \\

			\chatcompanion{} & CORE    & \xmark                      & 7.06                  & 14.77                 & 8.08                   \\
			\chatcompanion{} & CORE    & \cmark                      & 8.93                  & 33.87                 & 11.80                  \\
			\bottomrule
		\end{tabular}
	}
		    
	\caption{Evaluation results on Wiki QA. Each model retrieves top-5 knowledge snippets from the corresponding knowledge source. The top-5 answer recall of consolidated evidence (CORE) is $50.83$.}
	\label{tab:ottqa_results}
\end{table*}

\subsection{Dataset}
\paragraph{OTT-QA:} The OTT-QA dataset is an open-domain question answering benchmark that considers multi-step joint reasoning over both tabular and textual information. It consists of around 40K instances built upon Wikipedia, including 400K tables and 6M passages as the knowledge source.
Solving the questions in OTT-QA requires diverse reasoning skills and can be divided into three categories: single-hop questions (13\%), two-hop questions (57\%), and multi-hop questions (30\%). In this paper, we denote the dataset as Wiki QA.

\subsection{Experiment Setups}
In the following, we describe the experimental setup for Wiki QA.
Unless specified otherwise, the setups are identical to those used in Section~\ref{sec:exp_kgr}.
\paragraph{Knowledge Consolidator:}  Here, the Knowledge Consolidator uses Wikipedia passages and tables as the knowledge source.
Instead of using BM25 as done for dialog tasks, we resort to a dense model, DPR \cite{karpukhin-etal-2020-dense}, as the backbone retriever.
For DPR, both question and passage/table inputs are represented by the corresponding special token [CLS] embeddings from their respective encoders, and retrieval is simply done via maximum inner product search in the vector space. 
Given a question, we use DPR to obtain the initial set of evidence, which includes tables and passages.
As most WikiQA questions require reasoning hops across different pieces of information (\eg hopping from the album table to its entry artist page in \autoref{fig:motivation}), we contend that directly feeding this raw evidence set to Working Memory is insufficient for prompting LLMs.
Thus, we further use additional intermediary modules, \ie linker and chainer, from CORE \cite{ma2022open} to consolidate the raw evidence, including connecting relevant documents, reranking evidence, and splicing them into evidence chains.
We refer to \citet{ma2022open} for more details.

\paragraph{Prompt Engine:} The prompt templates utilized for Wiki QA is shown in the appendix in Table \ref{tab:appendix_prompt_templates_wikiqa}.

\paragraph{Utility:} 
Here, as a response to a given question is deemed to leverage information from the consolidated knowledge, we use recall as the utility score, \ie preferring responses with higher token overlap with the corresponding evidence set.
Similar to Section~\ref{sec:exp_kgr}, we again consider a template-based natural language generator for giving feedback to ChatGPT.

\paragraph{Evaluation Metrics:} As WikiQA mainly concerns short-form answers, we evaluate the generated responses using the token-level precision, recall and F1 scores against the annotated answers.

\subsection{Results}
Table \ref{tab:ottqa_results} presents the evaluation results on Wiki QA. 
As expected, the closed-book model alone performs very poorly. 
Based on our manual inspections, we find that most error cases are hallucinated answers and ChatGPT abstains from answering for $17\%$ cases.
We observe that incorporating knowledge obtained from either DPR or CORE significantly improves the F1 score. The substantial improvements observed over the closed-book ChatGPT model indicate the importance of enhancing LLMs with external knowledge.
Compared with raw evidence from DPR (row 2), we observe that consolidated evidence from our proposed Knowledge Consolidator with CORE (row 3) is more useful to the frozen ChatGPT model, achieving more pronounced improvements across the board.
This suggests that it is crucial to consolidate knowledge for eliciting black-box LLMs to perform grounded reasoning.
Lastly, consistent with the observations for news chat and customer service scenarios in Section~\ref{sec:exp_kgr}, augmenting ChatGPT with automated feedback further improves alignments (adapting ChatGPT to perform multi-step grounded reasoning), leading to a substantial increase in recall and F1 scores.

Compared with the state-of-the-art fine-tuned model \cite{ma2022open} using top-50 consolidated evidence, there still remains a noticeable gap in performance.
Besides a lower answer recall of the consolidated evidence, we attribute it to extra alignments required for ChatGPT to respond in a more concise way and conduct faithful step-by-step reasoning.
Therefore, there is ample room for future explorations on elicitive prompting to achieve further improvements.

\section{Related Work}
\label{sec:relatedwork}
Numerous LLMs for text generation
\cite{Radford2018} have been proposed over the years, including very competitive ones such as GPT-3 \cite{NEURIPS2020_1457c0d6,Ouyang2022TrainingLM},
OPT \cite{Zhang2022OPTOP},
GPT-j \cite{GPT-j}, and ChatGPT. However, most of them do not naturally incorporate external knowledge.
To address this limitation, various works augment LLMs with knowledge consisting of \eg 
personalized recommendations \cite{ghazvininejad2017knowledge},
Wikipedia article and web search \cite{dinan2018wizard,Shuster2022BlenderBot3A},
structured and unstructured knowledge of task-oriented dialog \cite{Peng2022GODELLP}.
Recent advances have focused on jointly finetuning the retriever and generation components of retrieval-augmented text generation systems
\cite{Lewis2020RetrievalAugmentedGF,Zhang2021RetGenAJ}, but these methods are not applicable to black-box LLMs.

More recent work attempts to combine black-box LLMs with external knowledge, such as incorporating external knowledge into prompts \cite{madaan2022memory,lazaridou2022internet}, 
making GPT-3 more faithful \cite{he2022rethinking}, 
and combining web knowledge with GPT-3 \cite{nakano2021webgpt}.
In very recent works related to ours,
\citet{Shi2023REPLUGRB} tune the ranker of a black-box LLM.
\citet{Toolformer} tune black-box LLMs' access to different APIs and show improvement on a variety of understanding and reasoning tasks. 
We consider these works to complementary to ours, as we assume our set of APIs to be given and fixed, and we instead focus more on when and what APIs to request, interactive feedback with the LLM, and developing a self-learning ability through utility functions.

\section{Limitations and Future Directions}
\label{sec:limitation}
A main limitation of this work is that interactive feedback with a computationally expensive model such as ChatGPT can significantly slow down the user experience, as ChatGPT is often queried twice for a single response. However, we think this can translate into more choice for the user. For example, the initial ChatGPT response can be shown to the user as it is being decoded, and the user could then be informed that a more accurate response is available (depending on the utility function). Then, an impatient user can decide to ignore this option, while a user more mindful of response accuracy may decide to see the improved ChatGPT response. In task-oriented and high-stakes scenarios, we believe many users would prefer the slower but more accurate option.

The main results of the paper are with a policy designed manually, as due to the current high-demand for ChatGPT and its limited bandwidth. As reinforcement learning can be quite sample inefficient, we trained our policy using an LLM (T5-Base) we could easily query, and these RL experiments demonstrate the effectiveness of \lmcompanion. As ChatGPT becomes more available, we plan to update the paper with RL experiments involving ChatGPT. The current version of the paper does not include human evaluation, as the goal with our current utility function (KF1) shown we can make ChatGPT more grounded and our experiments suggest the responses of our best system are better at capturing the words of the (gold) knowledge. As we move to towards much utility functions such as safety, it will be important to add more fine-grained analyzes of the responses, and we will add human evaluation.
In future work, we also plan to leverage interactions with real users and user feedbacks to train \lmcompanion.

\section{Conclusions}
\label{sec:conclusions}
We introduced \lmcompanion, a framework for augmenting black-box LLMs (e.g., ChatGPT) with external knowledge and automated feedback.
The external knowledge provided as part of the LLM prompts helps generate more responses that are more grounded into external knowledge relevant to the current conversation. The automated feedback elicits the ``follow-up correction''  abilities of models such as ChatGPT and InstructGPT in order to produce revised responses that rank higher according to some given utility functions (e.g., groundedness as measured by KF1). 
These various components are integrated together as part of an RL framework, which we optimize end-to-end using policy gradient. End-to-end experiments with T5 show the effectiveness of \lmcompanion, while experiments on ChatGPT show significant increases both in terms of KF1 and a host of text generation metrics.

\section*{Ethics Statement}
\label{sec:ethics}
It is widely understood that large language models have the potential to generate harmful, offensive, and inappropriate content \cite{Bender2021OnTD,Bommasani2021OnTO,Weidinger2021EthicalAS}. This paper is an attempt to address a major harm of LLMs, namely factual integrity. This paper does not address the problem of offensive content generation, but future work on \lmcompanion could help mitigate such harm via, \eg offensiveness-related utility functions.

As with other knowledge-augmented text generation applications, we cannot rule out that external sources could compromise the factuality of generated text. It is, therefore, important to encourage users to check the relevance of external sources that supplement the generated text.

\section*{Acknowledgements}

We thank 
Saleema Amershi,
Ahmed Awadallah,
Nguyen Bach,
Paul Bennett,
Chris Brockett,
Weixin Cai,
Dhivya Eswaran,
Adam Fourney,
Hsiao-Wuen Hon, 
Chunyuan Li,
Ricky Loynd,
Hoifung Poon,
Corby Rosset,
Bin Yu, 
Sheng Zhang, 
and members of the Microsoft Research Deep Learning group for valuable discussions and comments.

\bibliography{references}

\begin{thebibliography}{43}
\expandafter\ifx\csname natexlab\endcsname\relax\def\natexlab#1{#1}\fi

\bibitem[{Banerjee and Lavie(2005)}]{banerjee2005meteor}
Satanjeev Banerjee and Alon Lavie. 2005.
\newblock {METEOR}: An automatic metric for mt evaluation with improved
  correlation with human judgments.
\newblock In \emph{ACL Workshop on Intrinsic and Extrinsic Evaluation Measures
  for Machine Translation and/or Summarization}, pages 65--72.

\bibitem[{Bender et~al.(2021)Bender, Gebru, McMillan-Major, and
  Shmitchell}]{Bender2021OnTD}
Emily~M. Bender, Timnit Gebru, Angelina McMillan-Major, and Shmargaret
  Shmitchell. 2021.
\newblock On the dangers of stochastic parrots: Can language models be too big?
\newblock \emph{Proceedings of the 2021 ACM Conference on Fairness,
  Accountability, and Transparency}.

\bibitem[{Bommasani et~al.(2021)Bommasani, Hudson, Adeli, Altman, Arora, von
  Arx, Bernstein, Bohg, Bosselut, Brunskill, Brynjolfsson, Buch, Card,
  Castellon, Chatterji, Chen, Creel, Davis, Demszky, Donahue, Doumbouya,
  Durmus, Ermon, Etchemendy, Ethayarajh, Fei-Fei, Finn, Gale, Gillespie, Goel,
  Goodman, Grossman, Guha, Hashimoto, Henderson, Hewitt, Ho, Hong, Hsu, Huang,
  Icard, Jain, Jurafsky, Kalluri, Karamcheti, Keeling, Khani, Khattab, Koh,
  Krass, Krishna, Kuditipudi, Kumar, Ladhak, Lee, Lee, Leskovec, Levent, Li,
  Li, Ma, Malik, Manning, Mirchandani, Mitchell, Munyikwa, Nair, Narayan,
  Narayanan, Newman, Nie, Niebles, Nilforoshan, Nyarko, Ogut, Orr,
  Papadimitriou, Park, Piech, Portelance, Potts, Raghunathan, Reich, Ren, Rong,
  Roohani, Ruiz, Ryan, R'e, Sadigh, Sagawa, Santhanam, Shih, Srinivasan,
  Tamkin, Taori, Thomas, Tram{\`e}r, Wang, Wang, Wu, Wu, Wu, Xie, Yasunaga,
  You, Zaharia, Zhang, Zhang, Zhang, Zhang, Zheng, Zhou, and
  Liang}]{Bommasani2021OnTO}
Rishi Bommasani, Drew~A. Hudson, Ehsan Adeli, Russ Altman, Simran Arora, Sydney
  von Arx, Michael~S. Bernstein, Jeannette Bohg, Antoine Bosselut, Emma
  Brunskill, Erik Brynjolfsson, S.~Buch, Dallas Card, Rodrigo Castellon,
  Niladri~S. Chatterji, Annie~S. Chen, Kathleen~A. Creel, Jared Davis, Dora
  Demszky, Chris Donahue, Moussa Doumbouya, Esin Durmus, Stefano Ermon, John
  Etchemendy, Kawin Ethayarajh, Li~Fei-Fei, Chelsea Finn, Trevor Gale,
  Lauren~E. Gillespie, Karan Goel, Noah~D. Goodman, Shelby Grossman, Neel Guha,
  Tatsunori Hashimoto, Peter Henderson, John Hewitt, Daniel~E. Ho, Jenny Hong,
  Kyle Hsu, Jing Huang, Thomas~F. Icard, Saahil Jain, Dan Jurafsky, Pratyusha
  Kalluri, Siddharth Karamcheti, Geoff Keeling, Fereshte Khani, O.~Khattab,
  Pang~Wei Koh, Mark~S. Krass, Ranjay Krishna, Rohith Kuditipudi, Ananya Kumar,
  Faisal Ladhak, Mina Lee, Tony Lee, Jure Leskovec, Isabelle Levent, Xiang~Lisa
  Li, Xuechen Li, Tengyu Ma, Ali Malik, Christopher~D. Manning, Suvir
  Mirchandani, Eric Mitchell, Zanele Munyikwa, Suraj Nair, Avanika Narayan,
  Deepak Narayanan, Benjamin Newman, Allen Nie, Juan~Carlos Niebles, Hamed
  Nilforoshan, J.~F. Nyarko, Giray Ogut, Laurel~J. Orr, Isabel Papadimitriou,
  Joon~Sung Park, Chris Piech, Eva Portelance, Christopher Potts, Aditi
  Raghunathan, Robert Reich, Hongyu Ren, Frieda Rong, Yusuf~H. Roohani, Camilo
  Ruiz, Jack Ryan, Christopher R'e, Dorsa Sadigh, Shiori Sagawa, Keshav
  Santhanam, Andy Shih, Krishna~Parasuram Srinivasan, Alex Tamkin, Rohan Taori,
  Armin~W. Thomas, Florian Tram{\`e}r, Rose~E. Wang, William Wang, Bohan Wu,
  Jiajun Wu, Yuhuai Wu, Sang~Michael Xie, Michihiro Yasunaga, Jiaxuan You,
  Matei~A. Zaharia, Michael Zhang, Tianyi Zhang, Xikun Zhang, Yuhui Zhang,
  Lucia Zheng, Kaitlyn Zhou, and Percy Liang. 2021.
\newblock On the opportunities and risks of foundation models.
\newblock \emph{ArXiv}, abs/2108.07258.

\bibitem[{Brown et~al.(2020)Brown, Mann, Ryder, Subbiah, Kaplan, Dhariwal,
  Neelakantan, Shyam, Sastry, Askell, Agarwal, Herbert-Voss, Krueger, Henighan,
  Child, Ramesh, Ziegler, Wu, Winter, Hesse, Chen, Sigler, Litwin, Gray, Chess,
  Clark, Berner, McCandlish, Radford, Sutskever, and
  Amodei}]{NEURIPS2020_1457c0d6}
Tom Brown, Benjamin Mann, Nick Ryder, Melanie Subbiah, Jared~D Kaplan, Prafulla
  Dhariwal, Arvind Neelakantan, Pranav Shyam, Girish Sastry, Amanda Askell,
  Sandhini Agarwal, Ariel Herbert-Voss, Gretchen Krueger, Tom Henighan, Rewon
  Child, Aditya Ramesh, Daniel Ziegler, Jeffrey Wu, Clemens Winter, Chris
  Hesse, Mark Chen, Eric Sigler, Mateusz Litwin, Scott Gray, Benjamin Chess,
  Jack Clark, Christopher Berner, Sam McCandlish, Alec Radford, Ilya Sutskever,
  and Dario Amodei. 2020.
\newblock Language models are few-shot learners.
\newblock In \emph{Advances in Neural Information Processing Systems},
  volume~33, pages 1877--1901.

\bibitem[{Chowdhery et~al.(2022)Chowdhery, Narang, Devlin, Bosma, Mishra,
  Roberts, Barham, Chung, Sutton, Gehrmann et~al.}]{chowdhery2022palm}
Aakanksha Chowdhery, Sharan Narang, Jacob Devlin, Maarten Bosma, Gaurav Mishra,
  Adam Roberts, Paul Barham, Hyung~Won Chung, Charles Sutton, Sebastian
  Gehrmann, et~al. 2022.
\newblock Palm: Scaling language modeling with pathways.
\newblock \emph{arXiv preprint arXiv:2204.02311}.

\bibitem[{Dinan et~al.(2018)Dinan, Roller, Shuster, Fan, Auli, and
  Weston}]{dinan2018wizard}
Emily Dinan, Stephen Roller, Kurt Shuster, Angela Fan, Michael Auli, and Jason
  Weston. 2018.
\newblock Wizard of {Wikipedia}: Knowledge-powered conversational agents.
\newblock In \emph{International Conference on Learning Representations}.

\bibitem[{Eric et~al.(2019)Eric, Goel, Paul, Kumar, Sethi, Ku, Goyal, Agarwal,
  Gao, and Hakkani-Tur}]{eric2019multiwoz}
Mihail Eric, Rahul Goel, Shachi Paul, Adarsh Kumar, Abhishek Sethi, Peter Ku,
  Anuj~Kumar Goyal, Sanchit Agarwal, Shuyang Gao, and Dilek Hakkani-Tur. 2019.
\newblock Multiwoz 2.1: A consolidated multi-domain dialogue dataset with state
  corrections and state tracking baselines.
\newblock \emph{arXiv preprint arXiv:1907.01669}.

\bibitem[{Galley et~al.(2019)Galley, Brockett, Gao, Gao, and
  Dolan}]{dstc7task2}
Michel Galley, Chris Brockett, Xiang Gao, Jianfeng Gao, and Bill Dolan. 2019.
\newblock Grounded response generation task at {DSTC7}.
\newblock In \emph{AAAI Dialog System Technology Challenges Workshop}.

\bibitem[{Gao et~al.(2019)Gao, Galley, and Li}]{GaoEtAl:2019}
Jianfeng Gao, Michel Galley, and Lihong Li. 2019.
\newblock Neural approaches to conversational {AI}.
\newblock \emph{Foundations and Trends in Information Retrieval},
  13(2-3):127--298.

\bibitem[{Gao et~al.(2022)Gao, Xiong, Bennett, and Craswell}]{gao2022neural}
Jianfeng Gao, Chenyan Xiong, Paul Bennett, and Nick Craswell. 2022.
\newblock Neural approaches to conversational information retrieval.
\newblock \emph{arXiv preprint arXiv:2201.05176}.

\bibitem[{Ghazvininejad et~al.(2017)Ghazvininejad, Brockett, Chang, Dolan, Gao,
  Yih, and Galley}]{ghazvininejad2017knowledge}
Marjan Ghazvininejad, Chris Brockett, Ming{-}Wei Chang, Bill Dolan, Jianfeng
  Gao, Wen{-}tau Yih, and Michel Galley. 2017.
\newblock A knowledge-grounded neural conversation model.
\newblock \emph{CoRR}, abs/1702.01932.

\bibitem[{Glaese et~al.(2022)Glaese, McAleese, Trkebacz, Aslanides, Firoiu,
  Ewalds, Rauh, Weidinger, Chadwick, Thacker, Campbell-Gillingham, Uesato,
  Huang, Comanescu, Yang, See, Dathathri, Greig, Chen, Fritz, Elias, Green,
  Mokr'a, Fernando, Wu, Foley, Young, Gabriel, Isaac, Mellor, Hassabis,
  Kavukcuoglu, Hendricks, and Irving}]{sparrow}
Amelia Glaese, Nathan McAleese, Maja Trkebacz, John Aslanides, Vlad Firoiu,
  Timo Ewalds, Maribeth Rauh, Laura Weidinger, Martin Chadwick, Phoebe Thacker,
  Lucy Campbell-Gillingham, Jonathan Uesato, Po-Sen Huang, Ramona Comanescu,
  Fan Yang, A.~See, Sumanth Dathathri, Rory Greig, Charlie Chen, Doug Fritz,
  Jaume~Sanchez Elias, Richard Green, Sovna Mokr'a, Nicholas Fernando, Boxi Wu,
  Rachel Foley, Susannah Young, Iason Gabriel, William~S. Isaac, John F.~J.
  Mellor, Demis Hassabis, Koray Kavukcuoglu, Lisa~Anne Hendricks, and Geoffrey
  Irving. 2022.
\newblock Improving alignment of dialogue agents via targeted human judgements.
\newblock \emph{ArXiv}, abs/2209.14375.

\bibitem[{Guu et~al.(2020)Guu, Lee, Tung, Pasupat, and Chang}]{Guu2020REALMRL}
Kelvin Guu, Kenton Lee, Zora Tung, Panupong Pasupat, and Ming-Wei Chang. 2020.
\newblock {REALM}: Retrieval-augmented language model pre-training.
\newblock \emph{ArXiv}, abs/2002.08909.

\bibitem[{He et~al.(2022)He, Zhang, and Roth}]{he2022rethinking}
Hangfeng He, Hongming Zhang, and Dan Roth. 2022.
\newblock Rethinking with retrieval: Faithful large language model inference.
\newblock \emph{arXiv preprint arXiv:2301.00303}.

\bibitem[{Karpukhin et~al.(2020)Karpukhin, Oguz, Min, Lewis, Wu, Edunov, Chen,
  and Yih}]{karpukhin-etal-2020-dense}
Vladimir Karpukhin, Barlas Oguz, Sewon Min, Patrick Lewis, Ledell Wu, Sergey
  Edunov, Danqi Chen, and Wen-tau Yih. 2020.
\newblock \href {https://doi.org/10.18653/v1/2020.emnlp-main.550} {Dense
  passage retrieval for open-domain question answering}.
\newblock In \emph{Proceedings of the 2020 Conference on Empirical Methods in
  Natural Language Processing (EMNLP)}, pages 6769--6781, Online. Association
  for Computational Linguistics.

\bibitem[{Kim et~al.(2020)Kim, Eric, Gopalakrishnan, Hedayatnia, Liu, and
  Hakkani-Tur}]{kim2020beyond}
Seokhwan Kim, Mihail Eric, Karthik Gopalakrishnan, Behnam Hedayatnia, Yang Liu,
  and Dilek Hakkani-Tur. 2020.
\newblock Beyond domain {APIs}: Task-oriented conversational modeling with
  unstructured knowledge access.
\newblock In \emph{Proceedings of the 21th Annual Meeting of the Special
  Interest Group on Discourse and Dialogue}, pages 278--289.

\bibitem[{Kim et~al.(2023)Kim, Gella, Jin, Papangelis, Hedayatnia, Liu, and
  Hakkani-T{\"u}r}]{kim2023}
Seokhwan Kim, Spandana Gella, Di~Jin, Alexandros Papangelis, Behnam Hedayatnia,
  Yang Liu, and Dilek Hakkani-T{\"u}r. 2023.
\newblock {DSTC11} track proposal: Task-oriented conversational modeling with
  subjective knowledge.
\newblock \url{https://github.com/alexa/dstc11-track5}.

\bibitem[{Lazaridou et~al.(2022)Lazaridou, Gribovskaya, Stokowiec, and
  Grigorev}]{lazaridou2022internet}
Angeliki Lazaridou, Elena Gribovskaya, Wojciech Stokowiec, and Nikolai
  Grigorev. 2022.
\newblock Internet-augmented language models through few-shot prompting for
  open-domain question answering.
\newblock \emph{arXiv preprint arXiv:2203.05115}.

\bibitem[{Lewis et~al.(2020)Lewis, Perez, Piktus, Petroni, Karpukhin, Goyal,
  Kuttler, Lewis, tau Yih, Rockt{\"a}schel, Riedel, and
  Kiela}]{Lewis2020RetrievalAugmentedGF}
Patrick Lewis, Ethan Perez, Aleksandara Piktus, Fabio Petroni, Vladimir
  Karpukhin, Naman Goyal, Heinrich Kuttler, Mike Lewis, Wen tau Yih, Tim
  Rockt{\"a}schel, Sebastian Riedel, and Douwe Kiela. 2020.
\newblock Retrieval-augmented generation for knowledge-intensive nlp tasks.
\newblock \emph{ArXiv}, abs/2005.11401.

\bibitem[{Lian et~al.(2019)Lian, Xie, Wang, Peng, and Wu}]{Lian2019LearningTS}
Rongzhong Lian, Min Xie, Fan Wang, Jinhua Peng, and Hua Wu. 2019.
\newblock Learning to select knowledge for response generation in dialog
  systems.
\newblock In \emph{International Joint Conference on Artificial Intelligence}.

\bibitem[{Lin(2004)}]{lin2004rouge}
Chin-Yew Lin. 2004.
\newblock {ROUGE}: A package for automatic evaluation of summaries.
\newblock In \emph{ACL workshop}, pages 74--81.

\bibitem[{Ma et~al.(2022)Ma, Cheng, Liu, Nyberg, and Gao}]{ma2022open}
Kaixin Ma, Hao Cheng, Xiaodong Liu, Eric Nyberg, and Jianfeng Gao. 2022.
\newblock Open-domain question answering via chain of reasoning over
  heterogeneous knowledge.
\newblock \emph{arXiv preprint arXiv:2210.12338}.

\bibitem[{Madaan et~al.(2022)Madaan, Tandon, Clark, and
  Yang}]{madaan2022memory}
Aman Madaan, Niket Tandon, Peter Clark, and Yiming Yang. 2022.
\newblock Memory-assisted prompt editing to improve {GPT-3} after deployment.
\newblock \emph{arXiv preprint arXiv:2201.06009}.

\bibitem[{Nakano et~al.(2021)Nakano, Hilton, Balaji, Wu, Ouyang, Kim, Hesse,
  Jain, Kosaraju, Saunders et~al.}]{nakano2021webgpt}
Reiichiro Nakano, Jacob Hilton, Suchir Balaji, Jeff Wu, Long Ouyang, Christina
  Kim, Christopher Hesse, Shantanu Jain, Vineet Kosaraju, William Saunders,
  et~al. 2021.
\newblock {WebGPT}: Browser-assisted question-answering with human feedback.
\newblock \emph{arXiv preprint arXiv:2112.09332}.

\bibitem[{Ouyang et~al.(2022)Ouyang, Wu, Jiang, Almeida, Wainwright, Mishkin,
  Zhang, Agarwal, Slama, Ray, Schulman, Hilton, Kelton, Miller, Simens, Askell,
  Welinder, Christiano, Leike, and Lowe}]{Ouyang2022TrainingLM}
Long Ouyang, Jeff Wu, Xu~Jiang, Diogo Almeida, Carroll~L. Wainwright, Pamela
  Mishkin, Chong Zhang, Sandhini Agarwal, Katarina Slama, Alex Ray, John
  Schulman, Jacob Hilton, Fraser Kelton, Luke~E. Miller, Maddie Simens, Amanda
  Askell, Peter Welinder, Paul~Francis Christiano, Jan Leike, and Ryan~J. Lowe.
  2022.
\newblock Training language models to follow instructions with human feedback.
\newblock \emph{ArXiv}, abs/2203.02155.

\bibitem[{Papineni et~al.(2002)Papineni, Roukos, Ward, and
  Zhu}]{papineni2002bleu}
Kishore Papineni, Salim Roukos, Todd Ward, and Wei-Jing Zhu. 2002.
\newblock {BLEU}: a method for automatic evaluation of machine translation.
\newblock In \emph{ACL}, pages 311--318.

\bibitem[{Peng et~al.(2022)Peng, Galley, He, Brockett, Lid{\'e}n, Nouri, Yu,
  Dolan, and Gao}]{Peng2022GODELLP}
Baolin Peng, Michel Galley, Pengcheng He, Chris Brockett, Lars Lid{\'e}n, Elnaz
  Nouri, Zhou Yu, Bill Dolan, and Jianfeng Gao. 2022.
\newblock {GODEL}: Large-scale pre-training for goal-directed dialog.
\newblock \emph{ArXiv}, abs/2206.11309.

\bibitem[{Popovi{\'c}(2015)}]{popovic-2015-chrf}
Maja Popovi{\'c}. 2015.
\newblock chr{F}: character n-gram {F}-score for automatic {MT} evaluation.
\newblock In \emph{Proceedings of the Tenth Workshop on Statistical Machine
  Translation}, pages 392--395, Lisbon, Portugal. Association for Computational
  Linguistics.

\bibitem[{Radford et~al.(2018)Radford, Narasimhan, Salimans, and
  Sutskever}]{Radford2018}
Alec Radford, Karthik Narasimhan, Tim Salimans, and Ilya Sutskever. 2018.
\newblock Improving language understanding by generative pre-training.

\bibitem[{Schick et~al.(2023)Schick, Dwivedi-Yu, Dessì, Raileanu, Lomeli,
  Zettlemoyer, Cancedda, and Scialom}]{Toolformer}
Timo Schick, Jane Dwivedi-Yu, Roberto Dessì, Roberta Raileanu, Maria Lomeli,
  Luke Zettlemoyer, Nicola Cancedda, and Thomas Scialom. 2023.
\newblock Toolformer: Language models can teach themselves to use tools.

\bibitem[{Sellam et~al.(2020)Sellam, Das, and Parikh}]{sellam-etal-2020-bleurt}
Thibault Sellam, Dipanjan Das, and Ankur Parikh. 2020.
\newblock {BLEURT}: Learning robust metrics for text generation.
\newblock In \emph{Proceedings of the 58th Annual Meeting of the Association
  for Computational Linguistics}, pages 7881--7892, Online.

\bibitem[{Shi et~al.(2023)Shi, Min, Yasunaga, Seo, James, Lewis, Zettlemoyer,
  and tau Yih}]{Shi2023REPLUGRB}
Weijia Shi, Sewon Min, Michihiro Yasunaga, Minjoon Seo, Rich James, Mike Lewis,
  Luke Zettlemoyer, and Wen tau Yih. 2023.
\newblock {REPLUG}: Retrieval-augmented black-box language models.
\newblock \emph{ArXiv}, abs/2301.12652.

\bibitem[{Shuster et~al.(2021)Shuster, Poff, Chen, Kiela, and
  Weston}]{shuster2021retrieval}
Kurt Shuster, Spencer Poff, Moya Chen, Douwe Kiela, and Jason Weston. 2021.
\newblock Retrieval augmentation reduces hallucination in conversation.
\newblock \emph{arXiv preprint arXiv:2104.07567}.

\bibitem[{Shuster et~al.(2022)Shuster, Xu, Komeili, Ju, Smith, Roller, Ung,
  Chen, Arora, Lane, Behrooz, Ngan, Poff, Goyal, Szlam, Boureau, Kambadur, and
  Weston}]{Shuster2022BlenderBot3A}
Kurt Shuster, Jing Xu, Mojtaba Komeili, Da~Ju, Eric~Michael Smith, Stephen
  Roller, Megan Ung, Moya Chen, Kushal Arora, Joshua Lane, Morteza Behrooz,
  W.K.F. Ngan, Spencer Poff, Naman Goyal, Arthur~D. Szlam, Y-Lan Boureau,
  Melanie Kambadur, and Jason Weston. 2022.
\newblock {BlenderBot} 3: a deployed conversational agent that continually
  learns to responsibly engage.
\newblock \emph{ArXiv}, abs/2208.03188.

\bibitem[{Wang and Komatsuzaki(2021)}]{GPT-j}
Ben Wang and Aran Komatsuzaki. 2021.
\newblock {GPT-J-6B}: A 6 billion parameter autoregressive language model.
\newblock \url{https://github.com/kingoflolz/mesh-transformer-jax}.

\bibitem[{Weidinger et~al.(2021)Weidinger, Mellor, Rauh, Griffin, Uesato,
  Huang, Cheng, Glaese, Balle, Kasirzadeh, Kenton, Brown, Hawkins, Stepleton,
  Biles, Birhane, Haas, Rimell, Hendricks, Isaac, Legassick, Irving, and
  Gabriel}]{Weidinger2021EthicalAS}
Laura Weidinger, John F.~J. Mellor, Maribeth Rauh, Conor Griffin, Jonathan
  Uesato, Po-Sen Huang, Myra Cheng, Mia Glaese, Borja Balle, Atoosa Kasirzadeh,
  Zachary Kenton, Sande~Minnich Brown, William~T. Hawkins, Tom Stepleton,
  Courtney Biles, Abeba Birhane, Julia Haas, Laura Rimell, Lisa~Anne Hendricks,
  William~S. Isaac, Sean Legassick, Geoffrey Irving, and Iason Gabriel. 2021.
\newblock Ethical and social risks of harm from language models.
\newblock \emph{ArXiv}, abs/2112.04359.

\bibitem[{Williams(1992)}]{Williams1992SimpleSG}
Ronald~J. Williams. 1992.
\newblock Simple statistical gradient-following algorithms for connectionist
  reinforcement learning.
\newblock \emph{Machine Learning}, 8:229--256.

\bibitem[{Yeh et~al.(2021)Yeh, Eskenazi, and Mehri}]{yeh2021}
Yi-Ting Yeh, Maxine Eskenazi, and Shikib Mehri. 2021.
\newblock A comprehensive assessment of dialog evaluation metrics.
\newblock In \emph{arXiv}.

\bibitem[{Yuan et~al.(2021)Yuan, Neubig, and Liu}]{yuan2021}
Weizhe Yuan, Graham Neubig, and Pengfei Liu. 2021.
\newblock {BARTScore}: Evaluating generated text as text generation.

\bibitem[{Zhang et~al.(2022)Zhang, Roller, Goyal, Artetxe, Chen, Chen, Dewan,
  Diab, Li, Lin, Mihaylov, Ott, Shleifer, Shuster, Simig, Koura, Sridhar, Wang,
  and Zettlemoyer}]{Zhang2022OPTOP}
Susan Zhang, Stephen Roller, Naman Goyal, Mikel Artetxe, Moya Chen, Shuohui
  Chen, Christopher Dewan, Mona Diab, Xian Li, Xi~Victoria Lin, Todor Mihaylov,
  Myle Ott, Sam Shleifer, Kurt Shuster, Daniel Simig, Punit~Singh Koura, Anjali
  Sridhar, Tianlu Wang, and Luke Zettlemoyer. 2022.
\newblock {OPT}: Open pre-trained transformer language models.
\newblock \emph{ArXiv}, abs/2205.01068.

\bibitem[{Zhang et~al.(2019)Zhang, Kishore, Wu, Weinberger, and
  Artzi}]{zhang2019}
Tianyi Zhang, Varsha Kishore, Felix Wu, Kilian~Q. Weinberger, and Yoav Artzi.
  2019.
\newblock {BERTScore}: Evaluating text generation with {BERT}.
\newblock \emph{CoRR}, abs/1904.09675.

\bibitem[{Zhang et~al.(2021)Zhang, Sun, Gao, Fang, Brockett, Galley, Gao, and
  Dolan}]{Zhang2021RetGenAJ}
Yizhe Zhang, Siqi Sun, Xiang Gao, Yuwei Fang, Chris Brockett, Michel Galley,
  Jianfeng Gao, and Bill Dolan. 2021.
\newblock {RetGen}: A joint framework for retrieval and grounded text
  generation modeling.
\newblock In \emph{AAAI Conference on Artificial Intelligence}.

\bibitem[{Zhong et~al.(2022)Zhong, Lei, and Chen}]{Zhong2022TrainingLM}
Zexuan Zhong, Tao Lei, and Danqi Chen. 2022.
\newblock Training language models with memory augmentation.
\newblock \emph{ArXiv}, abs/2205.12674.

\end{thebibliography}
\bibliographystyle{acl_natbib}

\appendix

\vfill\null
\section{Appendix}
\label{sec:appendix}
\begin{table*}[ht]
	\centering
		    
	\setlength{\tabcolsep}{1.0mm}{
		\begin{tabular}{l|p{10.5cm}}
  \toprule
  Instruction & I want you to act as a chatbot AI for travel planning. You need to answer customer's questions nicely.    \\
  \midrule
 Context & \textbf{User:} Hello, I am looking for a moderately priced restaurant on the north side of town.
 
    \textbf{Assistant:} I recommend golden wok. It is in the north part of town with a moderate price range.
    
    \textbf{User:} Yes, I am looking for a moderately priced restaurant in the north part of town.
    
    \textbf{Assistant:} I recommend golden wok. It is in the north part of town with a moderate price range.
    
    \textbf{User:} I don't want Golden Wok, is there anything else?,
    
    \textbf{Assistant:} Yes, your other option is The Nirala. It serves Indian food and is also located in the north part of town.
    
    \textbf{User:} Is their beer a good value for the money? \\
  \midrule
  ChatGPT Response &  I am sorry, I do not have information on the beer prices at The Nirala. It is best to contact the restaurant directly for more information. \\
  \midrule
  \midrule
  Working Memory &  

Review: They also have a modest drinks menu with some affordable prices, but I was more interested in the beautiful view from my window seat, which allowed me to watch the sun setting as I dined.

Q: Is alcohol served at this restaurant? 
A: Yes, alcohol is served at this restaurant.

Review: I found them to be overpriced and mediocre.

Review: The drinks ae priced well.

Review: Something that I disliked, however was the prices for their drinks. \\
\midrule

\lmcompanion{} Response & I am sorry, I do not have information on the value of their beer. I only have information on the location and price range of the restaurant. \\

\midrule 

Rule-based Feedback & The response is inconsistent with the knowledge. Please generate again? \\

\midrule

\lmcompanion{} Response & Some reviews mention that the drinks at the restaurant are priced well, while others mention that they found the prices to be overpriced and mediocre. \\
\midrule
Human Response & Unfortunately the reviews are split, half say they are overpriced and have say they are affordable. \\
   \bottomrule
		\end{tabular}
	}
	\caption{Prompt and example responses of different models for the customer service scenarios.}
	\label{tab:appendix_example}
\end{table*}

Table~\ref{tab:appendix_example} provides sample responses contrasting ChatGPT and \lmcompanion{}. First, we can see that ChatGPT fails to provide a response related to specific knowledge related to the user, \eg a local Indian restaurant. In the second part of the table, we show \lmcompanion's Working Memory, which highlights the richer information retrieved from external knowledge to help the underling LLM (\ie ChatGPT as well) generate more contentful responses. The first LLM response received by \lmcompanion is unfortunately not satisfactory, as the quality and specificity of LLM generation can be unpredictable. In this case, the Utility module has determined that the first response did not meet its criteria (\ie KF1 above a given threshold), and issues a feedback to the LLM module (\ie ``response is inconsistent with the knowledge''). The second response received by \lmcompanion is much more satisfactory according to the utility function, and therefore sent to the user.

\begin{table}[ht]
	\centering
		    
	\setlength{\tabcolsep}{1.0mm}{
		\begin{tabular}{p{1\linewidth}}
  \toprule

I want you to act as a chatbot. You need to answer user' questions nicely. \\
  Context: \\
  User: [$U_1$] \\
  Assistant: [$R_1$] \\
  $\cdots$ \\
  User: [$U_t$] \\
  Assistant: \hl{[$R_t$]} \\
  \midrule
  I want you to act as a chatbot. You will be presented with knowledge snippets. You need to answer user' questions nicely and accurately based on the knowledge snippets. \\
  Working Memory: [$M_t$] \\
  Context: \\
  User: [$U_1$] \\
  Assistant: [$R_1$] \\
  $\cdots$ \\
  User: [$U_t$] \\
  Assistant: \hl{[$R_t$]} \\
  \bottomrule
  
		\end{tabular}
	}
	\caption{Prompt Templates for News Chat. LLMs generated responses is highlighted with \hl{[ ]}.}
	\label{tab:appendix_prompt_templates_news_chat}
\end{table}

\begin{table}[ht]
	\centering
		    
	\setlength{\tabcolsep}{1.0mm}{
		\begin{tabular}{p{1\linewidth}}
  \toprule
  
  I want you to act as a chatbot AI for travel planning. You need to answer customer’s questions nicely. \\
  Context: \\
  User: [$U_1$] \\
  Assistant: [$R_1$] \\
  $\cdots$ \\
  User: [$U_t$] \\
  Assistant: \hl{[$R_t$]} \\
  \midrule
  I want you to act as a chatbot AI for travel planning. You will be presented with knowledge snippets. You need to answer customer's questions nicely and accurately based on the knowledge snippets. \\
  Working Memory: [$M_t$] \\
  Context: \\
  User: [$U_1$] \\
  Assistant: [$R_1$] \\
  $\cdots$ \\
  User: [$U_t$] \\
  Assistant: \hl{[$R_t$]} \\
  \bottomrule
  
		\end{tabular}
	}
	\caption{Prompt Templates for Customer Service. LLMs generated response are highlighted with \hl{[ ]}.}
	\label{tab:appendix_prompt_templates_customer_service}
\end{table}

\begin{table}[ht]
	\centering
		    
	\setlength{\tabcolsep}{1.0mm}{
		\begin{tabular}{p{1\linewidth}}
  \toprule
  
  I am a highly intelligent question answering bot that can answer questions. If you ask me a question that is rooted in truth, I will give you the answer. If you ask me a question that is nonsense, trickery, or has no clear answer, I will respond with ``Unknown". \\
  Question: [$Q$] \\
  Answer: \hl{[$A$]} \\
  \midrule
  I am a highly intelligent question answering bot, and can answer questions given some documents and tables. If you ask me a question that is rooted in truth, I will give you the answer. If you ask me a question that is nonsense, trickery, or has no clear answer, I will respond with ``Unknown".  \\

Working Memory: [$M$] \\
Question: [$Q$] \\
  Answer: \hl{[$A$]} \\
  \bottomrule
  
		\end{tabular}
	}
	\caption{Prompt Templates for Wiki QA. LLMs generated response are highlighted with \hl{[ ]}.}
	\label{tab:appendix_prompt_templates_wikiqa}
\end{table}

\end{document}